\documentclass[letterpaper, 10 pt, conference]{ieeeconf}

\IEEEoverridecommandlockouts
\usepackage[utf8]{inputenc}
\usepackage[T1]{fontenc}
\usepackage{graphicx}
\usepackage{amsmath}
\usepackage{amssymb}
\usepackage{booktabs}
\usepackage{cite}
\usepackage{xcolor}
\usepackage{bm}
\usepackage[hidelinks]{hyperref}
\usepackage[capitalize]{cleveref}
\usepackage{censor}
\usepackage{eso-pic}

\title{\LARGE \bf
Picking Bins Empty: A Hierarchical Hybrid Approach with Online Self-Learning of Grasp Points for Reliable Industrial Bin-Picking
}

\author{Florian Töper$^{**1,2}$, Samarth Kishor Yelvande$^{**1}$, Jan Niklas Ewertz$^{1,3}$, Rudolph Triebel$^{3,4}$, Peter Ohlhausen$^{5}$%
\thanks{$^{*}$ This work was supported by the Federal Ministry for Economic Affairs and Energy based on a resolution of the German Bundestag with DARP}%
\thanks{$^{**}$ Authors contributed equally to this work.}%
\thanks{$^{1}$ \raggedright Mercedes-Benz AG, Sindelfingen, Germany \tt\small florian.toeper@mercedes-benz.com}%
\thanks{$^{2}$ University of Stuttgart, Germany}%
\thanks{$^{3}$ Karlsruhe Institute of Technology (KIT), Germany}%
\thanks{$^{4}$ DLR Institute of Robotics and Mechatronics, Weßling, Germany}%
\thanks{$^{5}$ Fraunhofer Institute for Industrial Engineering (IAO)}%
}

\AddToShipoutPictureBG*{%
  \ifnum\value{page}=1
    \AtPageLowerLeft{%
      \raisebox{16pt}{%
        \makebox[\paperwidth]{%
          \parbox{0.92\paperwidth}{\centering\fontsize{6.5}{7.5}\selectfont
          \copyright~2026 IEEE. Personal use of this material is permitted. Permission from IEEE must be obtained for all other uses, in any current or future media, including reprinting/republishing this material for advertising or promotional purposes, creating new collective works, for resale or redistribution to servers or lists, or reuse of any copyrighted component of this work in other works.}%
        }%
      }%
    }%
  \fi
}

\begin{document}

\maketitle
\thispagestyle{empty}
\pagestyle{empty}

\crefname{equation}{}{}
\crefformat{equation}{(#2#1#3)}
\Crefname{equation}{Equation}{Equations}%

\begin{abstract}
Bin-picking is a cornerstone of modern manufacturing, yet achieving complete bin clearance without manual intervention remains a critical challenge. 
While model-based methods provide high precision, they frequently suffer from deadlocks when predefined grasps are occluded or perception fails. 
Labor-intensive fine-tuning of grasp points is commonly required to reach a satisfactory performance for new parts.
Model-free algorithms offer a more flexible alternative with ``out-of-the-box'' versatility but lack the reliability and repeatability required for production.
Unlike existing work, which treats the two techniques in isolation, we propose a four-tiered hierarchical hybrid approach to combine the best of both worlds.
A model-based pipeline serves as a robust backbone, while a model-free ``exploration agent'' resolves deadlock situations and discovers new grasp points.
This is supported by an online self-learning mechanism that uses gripper-stroke feedback and Wilson score intervals to autonomously rank grasp candidates, reducing manual commissioning effort. 
Validation on three automotive parts demonstrates that our method significantly outperforms a model-free baseline in grasp success rate while improving the bin clearance rate of the model-based baseline from 50.9\% to 100\% across all experiments. 
This transition to full bin clearance marks a significant step towards truly autonomous, intervention-free industrial operation.
\end{abstract}

\section{INTRODUCTION}

\begin{figure}[!t]
    \centering
    \includegraphics[width=\columnwidth]{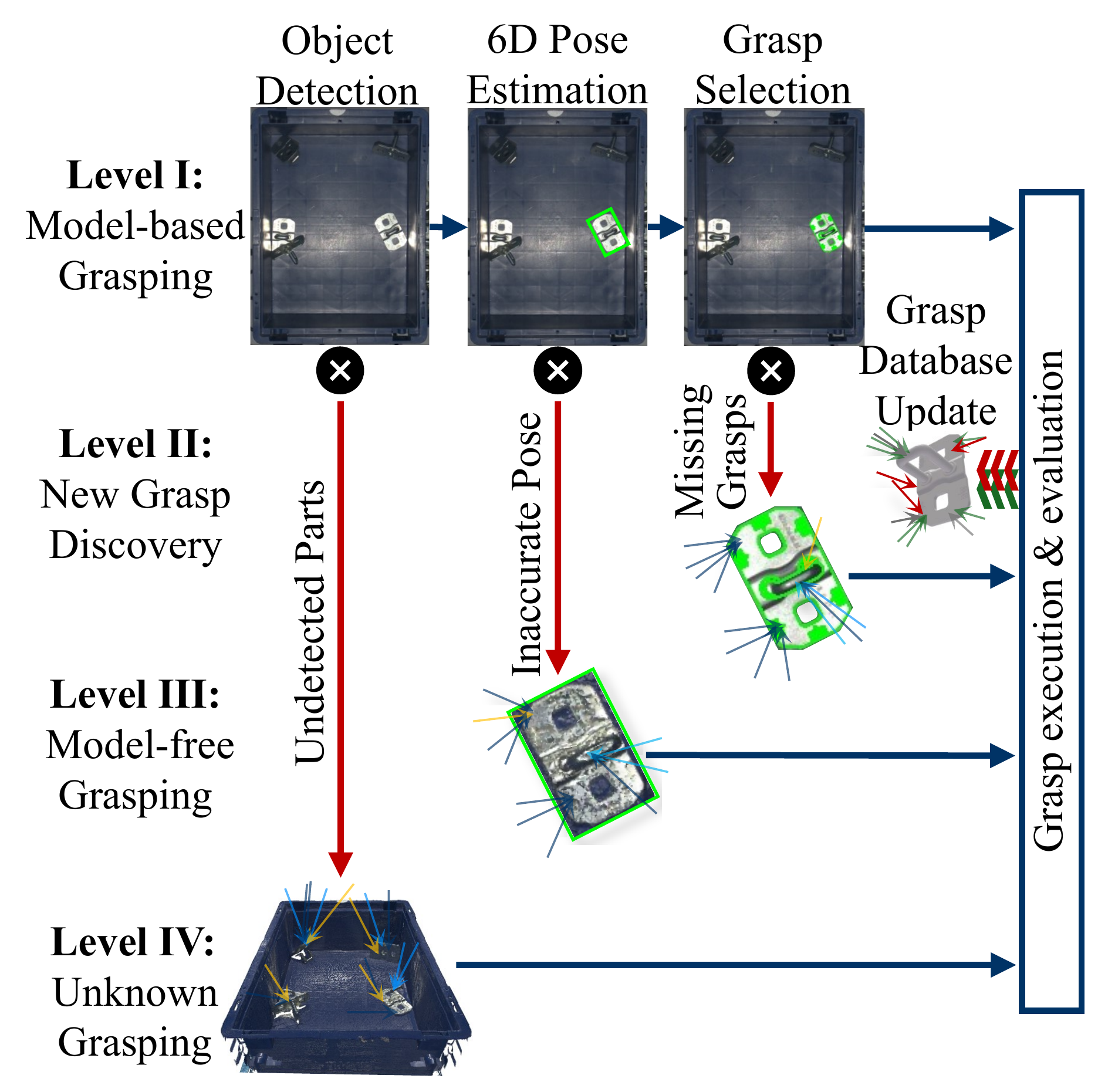}
    \caption{Our hierarchical system architecture with a model-based backbone on Level I and model-free grasp generation on Levels II to IV to achieve full bin clearance despite three potential points of failure. A self-learning mechanism on Levels I and II discovers and autonomously evaluates grasp candidates during operation to optimize the grasp success rate.}
    \label{fig:system_architecture}
\end{figure}

Robotic grasping in unstructured scenarios like random bin-picking is a key requirement for flexible manufacturing automation.
Recent years have shown a surge in autonomous robotics driven by ever-increasing computer vision capabilities.
Despite these algorithmic leaps, the implementation of bin-picking in real-world industrial operation remains limited~\cite{huang2025xyz}.
A key barrier is the common inability of current solutions to reliably empty bins without manual intervention or redesigned forms of material provision ~\cite{Liu2024EfficientED, spenrath2022heuristisches, sarna2023impact, Boschetti2023}. 
Industry demands autonomous operation for extended periods, yet existing solutions often encounter deadlock situations, particularly as bins reach a sparse state~\cite{spenrath2022heuristisches, Boschetti2023,Fujita2019}.

State-of-the-art industrial systems mostly rely on model-based methods, which utilize known CAD geometries for pose estimation and typically manual grasp point definition~\cite{pantano2024simplifying}. 
This provides higher robustness and repeatability and enables oriented placing for downstream tasks~\cite{pantano2024simplifying}. 

Yet, model-based bin-picking typically involves the labor-intensive task of manually defining---and often iteratively adjusting---grasp points until a satisfactory system performance is achieved~\cite{pantano2024simplifying, kleeberger2021automatic}. 
This increases reliance on experts and hinders the integration of diverse part catalogs, required for industrial use cases such as order picking. 
In addition, any perception failures from object detection, pose estimation, or missing annotated grasps directly lead to grasping errors and objects remaining in the bin~\cite{fu2024low}.
This limits the autonomy and throughput and thus affects the economic potential of bin-picking applications~\cite{spenrath2022heuristisches, Boschetti2023}.

Model-free methods, on the other hand, predict grasps directly from sensor data and promise object-agnostic “out-of-the-box” versatility~\cite{sundermeyer2021contact}. 
However, they often exhibit a lower grasp success rate (GSR), especially in cluttered industrial bins, where contact physics are complex~\cite{back2025graspclutter6d}. 
Also, they lack semantic understanding of the grasped objects, which limits repeatability and disallows oriented placing~\cite{khalid2021automatic}. 

This creates a critical trade-off: practitioners must choose between labor-intensive but more reliable model-based systems and more flexible but inconsistent model-free methods.

In contrast to this existing paradigm, we suggest treating the two methods not as distinct alternatives but rather as complementary techniques in a unified grasping pipeline. 
We propose a hierarchical hybrid approach with a robust model-based backbone that seamlessly falls back to a model-free backup and “exploration agent” in case of perception errors or missing grasp points.

We further introduce an online self-learning mechanism. 
By utilizing gripper-stroke feedback as an autonomous labeling signal and Wilson score intervals, the system autonomously discovers and ranks grasps based on historical success. 
This allows the robot to learn effective and ineffective grasps during operation, effectively commissioning itself without human oversight.

Our contribution can be summarized as follows:
\begin{itemize}
    \item \textbf{Hierarchical Fallback Strategy:} A four-tiered framework bridging 6D pose estimation and model-free grasping to enable 100\% bin clearance even under perception failures.
    \item \textbf{Online Self-learning Mechanism:} Autonomous grasp database expansion via model-free exploration of new grasp candidates, removing the need for human-in-the-loop parameter tuning.
    \item \textbf{Grasp Ranking Strategy:} Evaluation of grasp points using gripper-stroke feedback and Wilson score intervals to prune unsuccessful candidates and optimize grasping performance.
    \item \textbf{Industrial Validation:} Real-world benchmarking on three automotive parts, demonstrating reliable bin clearance in contrast to a model-based baseline.
\end{itemize}

\section{RELATED WORK}

Bin-picking combines sensing, perception, and actuation to autonomously pick \& place parts from (unstructured) load carriers~\cite{cordeiro2025review}. 
Methods can be divided into model-based and model-free approaches~\cite{kleeberger2020transferring, kleeberger2020survey}. 
Some systems employ hybrid approaches or online self-learning strategies.

\subsection{Model-Based Bin-Picking}

CAD models are commonly available in industry. 
Model-based (analytical) approaches utilize this geometric information for the two related challenges of localizing workpieces in a bin and selecting suitable grasp points~\cite{kleeberger2020transferring, moosmann2020increasing}.

State-of-the-art pipelines typically perform object detection and 6D pose estimation to accurately locate objects~\cite{huang2025lessons}.%
Recent years have shown remarkable advancements as showcased by the Benchmark for 6D Object Pose Estimation (BOP)~\cite{bop_challenges_website}. 
Still, challenges remain, especially for scenes with high clutter, occlusion, and reflective objects, which are common in industrial bin-picking~\cite{huang2025xyz}.

Given a 6D pose estimate, grasps can be predefined in an object coordinate system and checked for collision-free access and kinematic feasibility~\cite{kleeberger2020transferring}. 
This allows repeatable and precise grasping as well as oriented placing~\cite{kleeberger2020transferring}. 
Analytical methods therefore remain the dominant approach for industrial bin-picking~\cite{pantano2024simplifying}. 

The described workflow, however, inherently leads to three potential points of perception failure that limit the performance of existing bin-picking systems:
\begin{enumerate}
    \item False negatives in object detection lead to remaining undetected parts in the bin.
    \item Inaccurate pose estimates directly propagate to imprecise grasps, affecting the grasp success rate.
    \item Deadlocks occur if none of the predefined grasps of the located parts is collision-free and reachable.
\end{enumerate}

Consequently, error rates are still high in practical industrial applications.
Particularly for parallel-jaw grippers, full bin clearance remains difficult, especially in narrow spaces towards the corner of bins.
Errors are often resolved via manual intervention, negatively affecting throughput and economic potential.~\cite{spenrath2022heuristisches, Boschetti2023}

Another main drawback is the commissioning effort for integrating new parts. 
In addition to setting up a reliable pose estimator, grasp poses must be selected, prioritized, and tuned until a satisfactory system performance is achieved~\cite{kleeberger2020transferring}. 
This limits the applicability in High-Mix Low-Volume (HMLV) manufacturing and increases reliance on experts~\cite{pantano2024simplifying, kleeberger2020transferring}.

For localization, recent research aims to reduce the burden of model training through the use of synthetic data~\cite{cao20236impose} or the application of zero-shot models~\cite{huang2025lessons, chen2025zerobp}.%
This shows promising results towards faster deployment but can come at the cost of increased errors due to differences between (synthetic) training data and real applications~\cite{hagelskjaer2025good}.

For grasping, previous work intends to overcome manual annotation by automatic grasp pose synthesis from the workpiece geometry. 
Kleeberger et al.~\cite{kleeberger2021automatic} propose an offline approach for parallel grippers using features like gripper stroke, surface friction, and a gripper collision check to determine grasp candidates. 
They apply a clustering algorithm to limit the number of grasps while maintaining high variety.
Khalid et al. present a similar approach for suction grippers~\cite{khalid2021automatic}.

As geometric features alone do not accurately reflect the physical interaction between gripper and object, subsequent work extends automatic grasp pose synthesis by physics simulation~\cite{kleeberger2020transferring}. 
While this boosts grasp success rates, it creates additional overhead, and a sim-to-real gap remains, especially for the complex contact dynamics in real cluttered scenes~\cite{kleeberger2020transferring, back2025graspclutter6d}. 
This issue is further exacerbated in case form-flexible grippers are used.

\subsection{Model-Free Bin-Picking}

In contrast to model-based methods, model-free techniques predict grasp hypotheses directly from sensor data, eliminating the need for pose estimation and object-specific grasp annotation~\cite{cordeiro2025review, kleeberger2020survey}. 
State-of-the-art methods employ deep learning (DL) to estimate collision-free grasps from depth data~\cite{pantano2024simplifying}.
(Unknown) object instance segmentation is often used to specify a target object~\cite{sundermeyer2021contact, murali2025graspgen}.

Discriminative algorithms select grasp points by evaluating the quality of sampled grasp candidates~\cite{kleeberger2020survey, sundermeyer2021contact}.
For instance, Dex-Net 1.0~\cite{mahler2016dex} matches current sensor observations against a database of known successful grasps for more than 10,000 object models.
A prior belief distribution is calculated using depth data, grasp parameters, and DL-based object similarity estimation to ultimately execute the candidate with the maximum lower confidence bound~\cite{mahler2016dex}.

Generative methods instead directly synthesize grasp poses and graspability scores~\cite{kleeberger2020survey, sundermeyer2021contact, murali2025graspgen}.
Contact-GraspNet (CGN)~\cite{sundermeyer2021contact} predicts 6-DoF parallel-jaw grasps from 3D point clouds end-to-end, treating sampled surface points as potential grasp contacts. 
This anchoring limits the predicted grasp representation to the gripper width and rotation, facilitating the learning process~\cite{sundermeyer2021contact}.

More recently, GraspGen~\cite{murali2025graspgen} employs 6-DoF grasp generation by iterative diffusion for parallel and suction grippers, paired with a learned discriminator to evaluate grasps. 
While the method shows superior success rates on benchmarks compared to prior approaches, it is computationally demanding and can predict grasps unbound to the object surface in contrast to contact-point architectures like CGN~\cite{murali2025graspgen}.

The successful generalization of all learning-based methods relies on large and diverse training datasets that closely reflect the target distribution. 
GraspClutter6D~\cite{back2025graspclutter6d} provides a large-scale real-world dataset covering 75 different objects in dense arrangements.
The authors show a significant performance increase for CGN, particularly for cluttered scenes, compared to previous training datasets consisting of simplistic scenes and synthetic data.~\cite{sundermeyer2021contact}

Despite significant progress, learning-based grasping approaches have yet to achieve robust real-world deployment across different robot embodiments and grippers~\cite{murali2025graspgen}. 
GraspGen achieves 83.3\% for simple cluttered tabletop scenes, whereas CGN achieves 67.9\% in more complex scenes of 15 piled objects when trained on GraspClutter6D~\cite{back2025graspclutter6d}~\cite{murali2025graspgen}. 

With that, model-free approaches remain inferior in industrial use cases with a single known object type per bin~\cite{pantano2024simplifying, moosmann2020increasing}. 
Besides, they are less deterministic and lack the geometric domain knowledge required for oriented placing or the detection of entanglements~\cite{moosmann2020increasing}. 
Lastly, they create predictions without learning from experience, which means being bound to repeat the same mistakes repeatedly~\cite{Sanders2020}.

\subsection{Hybrid Approaches and Online Self-Learning}

For both model-based and model-free approaches, failed grasps remain a common occurrence. 
Whereas Kleeberger et al.~\cite{kleeberger2020transferring} focus on offline simulation to improve grasp success, other research focuses on hybrid strategies or self-supervision in real operation to avoid the sim-to-real gap.

Examples include the self-supervised collection of more than 50k grasps with a parallel gripper by Pinto and Gupta~\cite{pinto2016supersizing} and over 800k grasps by Levine et al.~\cite{levine2018learning}. 
Similar trial-and-error approaches use the sensor signal of a vacuum gripper to provide labels for each grasp~\cite{monorchio2018learning}.

However, collecting real-world data to learn grasping policies from scratch is time-consuming, especially for reinforcement learning methods, and sensitive to modifications in the setup~\cite{kleeberger2020survey, patten2020dgcm}. 
It can thus be more practical to limit self-learning to the gradual improvement of an existing baseline. 
Hagelskjær~\cite{hagelskjaer2025good} applies this idea to fine-tune a 6D pose estimator in operation using in-hand pose estimation of grasped objects for verification.

Komoda et al.~\cite{komoda2024hybrid} describe a hybrid setup where a rule-based module explicitly models geometrical and physical constraints for high grasp success. 
A DL model is trained on the outputs of the rule-based system, allowing faster inference~\cite{komoda2024hybrid}. 
Dynamic selection between the models maintains high grasp success while improving the throughput~\cite{komoda2024hybrid}.
This exploitation of complementary strengths follows a similar thought as our proposed solution. 
However, the DL model is trained entirely from rule-based outputs, capping performance at the quality of handcrafted rules, rather than extending beyond them. 

In summary, existing approaches mostly focus on offline pre-computation or geometric generalization.
In contrast, our method aims at a dynamic, part-specific database expansion during live operation to resolve unavoidable deadlock situations.
It estimates the actual success rates of precise grasp poses based on real-world experience to prioritize the most robust candidates and prune unsuccessful grasps in real-time.
To fulfill the strict requirements of industrial bin-picking, we include a 6D pose estimation step, which is essential for repeatable and high-precision grasping beyond the commonly evaluated everyday items in minimal clutter.

\section{METHODOLOGY}

\subsection{System Architecture}

Our setup consists of four levels as shown in \cref{fig:system_architecture}. 
On \textbf{Level I}, a model-based pipeline with an object detector, a 6D pose estimator, and a grasp database serves as the backbone of operation. 
The subsequent hierarchy levels tackle the three previously described potential points of perception failure in model-based systems, ensuing in inverse order.

\textbf{Level II} is executed if poses are estimated with high confidence but none of the grasps in the database are collision-free and reachable. 
A model-free “exploration agent” then generates and scores new grasp hypotheses and selects the best candidate to overcome deadlock situations. 
The existence of a reliable 6D pose estimate makes it possible to add the discovered grasp points to the database.
Grasp outcomes are autonomously evaluated using gripper-stroke feedback, forming the basis of an efficient online self-learning mechanism using Wilson score intervals.

\textbf{Level III} is reached if a part was successfully detected but the predicted 6D pose falls below a minimum confidence score. 
As inaccurate pose estimates lead to imprecise model-based grasps, model-free grasp planning takes over to avoid grasping failures. 
The available detection result is used to narrow down the target region in the 3D point cloud.

\textbf{Level IV} tackles errors in the initial detection step. 
For industrial bin-picking with single-variety bulk material, the assumption is that all objects in the bin correspond to the target object.
In that case, model-free grasping can be performed for the remaining 3D points within the bin. 
This level can be skipped if there is a high likelihood of foreign objects in the bins or extra packaging material is used.

\subsection{Model-Based Backbone}

Object detection and 6D pose estimation of a commercial bin-picking system~\cite{roboception_cadmatch} are used for a state-of-the-art backbone on Level I to allow the repeatable execution of high-precision grasps. 
Training is carried out for each part using fully synthetic training data with heavy domain randomization to enable fast integration of different parts.

The detector generates oriented bounding boxes along with a confidence score for each prediction. 
Predictions above a certain confidence threshold are passed on to the 6D pose estimation module. 
Object poses are refined using depth data, and a fitness score is calculated by matching the projected CAD model with the 3D point cloud.

For remaining high-fitness poses after thresholding, existing grasps in a database are checked for collision-free access of the gripper, considering the observed point cloud, the detected load carrier, and other identified parts. 
The top-ranked collision-free grasp amongst all detected parts is executed.
The grasp database is either initialized manually via a user interface or, alternatively, set off empty, relying only on the model-free exploration of grasp candidates.

\subsection{Model-Free Fallback and Grasp Exploration}
We extend the pipeline of Level I by fallback strategies on Levels II, III, and IV using the grasp generation network Contact-GraspNet, trained on GraspClutter6D to enhance the real-world grasping performance in cluttered scenes~\cite{sundermeyer2021contact, back2025graspclutter6d}. 
The contact-point architecture of CGN ensures physical interaction with the object surface for each grasp.

As input for CGN, 2,048 points are randomly sampled from a target region along with 20,000 points of the surrounding context, with the load carrier and all detected object poses projected into the scene. 
For each target point, CGN predicts the gripper orientation and width along with a normalized confidence score for grasp success between zero and one.
As our gripper jaws are symmetric, we add an equivalent replication for each predicted grasp by flipping it 180° around its z-axis.

The target region for grasp generation is determined based on the available knowledge on each level. 
On Level II, the estimated 6D pose is projected into the point cloud, and points within 1 mm are selected to segment the target object. 
For Level III, the point cloud is cropped based on the oriented bounding box. 
On Level IV, grasps are predicted on the remaining 3D points inside the load carrier.

Level II serves the additional purpose of grasp exploration.
As the 6D object pose is known, selected grasp points can be added to expand the grasp database. 
To guide the system towards quickly learning reliable grasps, we extend the native grasp quality score predicted by CGN ($S_{CGN}$) by geometry-based scoring criteria for Level II. 
For this, a score, with empirically defined weights $\alpha$, $\beta$, and $\gamma$, is calculated as

\begin{equation}
    S_{total} = \alpha \cdot S_{CGN} + \beta \cdot S_{AP} + \gamma \cdot S_{COG} .
    \label{eq:s_total}
\end{equation}

$S_{AP}$ evaluates how antipodal the surface normals of the two expected contact points are. 
For antipodal grasps, the connecting vector $\mathbf{v}$ between the two surface points lies within the friction cones formed around the two surface normals $\mathbf{n_1}$ and $\mathbf{n_2}$~\cite{murray2017mathematical}. 
The half-angle of the friction cones $\phi$ is the arctan of the friction coefficient $\mu$~\cite{murray2017mathematical}. 
Subsequently, $S_{AP}$ can be calculated as

\begin{equation}
    S_{AP} = \max\!\left(0, \frac{\phi - \max\!\left(\angle\!\left(-\mathbf{n}_1,\mathbf{v}\right),\, \angle\!\left(\mathbf{n}_2,\mathbf{v}\right)\right)}{\phi + \epsilon}\right),
    \label{eq:s_ap}
\end{equation}

with $\epsilon$ as a small constant for numerical stability. 
As we use a flexible material on the gripper surface, we choose a high value of 0.9 for $\mu$.

Lastly, $S_{COG}$ describes the offset of a grasp with respect to the center of gravity, as rotational torque during lifting can affect the grasp stability. 
The score is calculated by \cref{eq:s_cog}, based on the offset $d$ to the estimated center of mass relative to the object's bounding-box diagonal $D_{obj}$.

\begin{equation}
    S_{COG} = 1 - \frac{2 \cdot d}{D_{obj}} 
    \label{eq:s_cog}
\end{equation}

\begin{figure}[t!]
    \centering
    \includegraphics[width=\columnwidth]{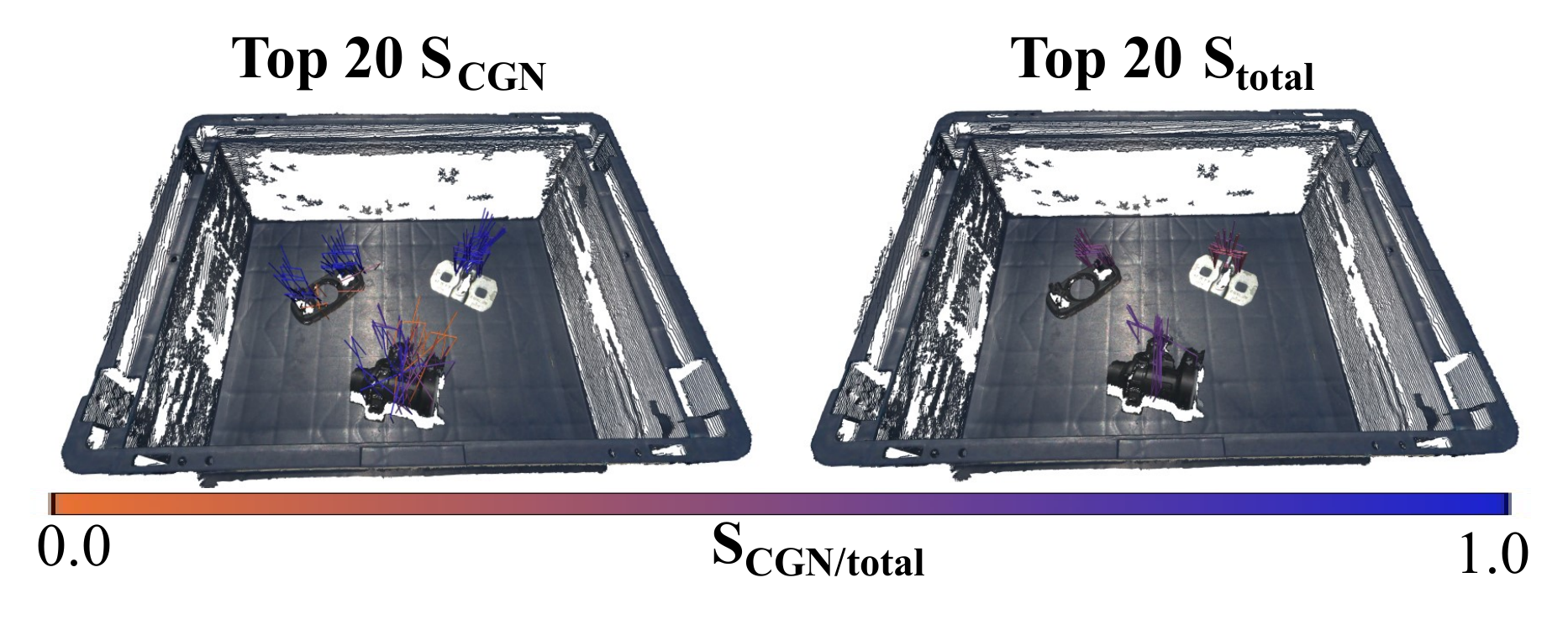}
    \caption{Effect of the reranking step on the top twenty grasp points predicted by CGN for A, B and C in an exemplary scene.}
    \label{fig:reranking}
\end{figure}

For our tests, we heuristically set $\alpha = 0.2$, $\beta = 0.6$, and $\gamma = 0.2$, as prior research highlights the importance of antipodal contact points for physical grasp stability~\cite{kleeberger2021automatic}. 
\cref{fig:reranking} shows the qualitative effect of the reranking step on the top twenty grasp points predicted by CGN for the parts A, B and C in an exemplary scene.

\subsection{Online Self-Learning and Ranking Logic}

Despite the reranking, the actual interaction of part and gripper remains difficult to evaluate analytically. 
Different grasp points can result in drastically different failure rates and varying tolerance to perception inaccuracy. 
Hence, we employ an online self-learning mechanism to iteratively evaluate grasp points based on real-world experience. 
The selection of high-scoring grasps enhances grasp success rates over time, maximizing the system’s throughput.

An electrical gripper allows us to accurately read the grasp width after each attempt to determine if a part slipped out. 
The number of executed grasps $n$ and successful attempts $k$ is updated for the specific grasp point after each attempt. 
A Wilson score interval 

\begin{equation}
    p_{o,u} = \frac{1}{1 + \frac{z^2}{n}} \cdot \left( \hat{p} + \frac{z^2}{2n} \pm z \cdot \sqrt{\frac{\hat{p} \cdot (1 - \hat{p})}{n} + \frac{z^2}{4n^2}} \right),
    \label{eq:wilson}
\end{equation}

with $\hat{p} = \frac{k}{n}$ is calculated to reflect the confidence that the next grasp attempt will be successful~\cite{Wilson01061927}.
We set $z$ to 1.96, equivalent to a 95\% confidence interval between $p_u$ and $p_o$.
For initial grasps with $n = 0$, we set $p_u$ to 0 and $p_o$ to 1.

Grasps are ranked by the upper border of the confidence interval $p_o$.
If two grasps have the identical value of $p_o$, we define a tiebreaker 

\begin{equation}
    t = \begin{cases} 
        k + 2 & \text{if predefined} \\ 
        k & \text{if discovered} 
    \end{cases}.
    \label{eq:tiebreaker}
\end{equation}

Finally, if the best grasp is collision-free for multiple parts, it is executed for the one with the highest pose fitness.
Ranking by $p_o$ ensures optimism in the face of uncertainty.
Untested grasps and those with a flawless history ($\hat{p} = 1$) achieve the maximum score $p_o = 1$.
The tiebreaker then dictates a greedy exploitation among these top candidates, favoring grasps that have proven their 100\% success rate most frequently.
Simultaneously, it acts as a soft prior to initially prefer predefined grasps while letting learned grasps quickly overtake existing ones if they prove to be more successful in practice. 
If no flawless grasp exists, untested grasp candidates ($p_o = 1$) are explored before the system retreats to grasps with a success rate below 100\% ($p_o < 1$).

We intentionally diverge from standard Upper Confidence Bound (UCB) algorithms~\cite{auer2002finite} and stochastic methods like Thompson sampling~\cite{daniel2018tutorial} 
as they force the eventual exploration of all options and thus risk the short-term selection of sub-optimal actions.
Our formulation reduces this risk by strictly prioritizing proven reliability to maintain a high GSR, while safely exploring new candidates when necessary.

For grasps with $p_o$ below 75\%, we disable further execution on Level I and the learning of similar grasps on Level II.
The 75\% threshold is chosen to instantly block grasps if the initial two grasping attempts are unsuccessful.
This hinders a single statistical outlier from instantly blocking a potentially viable grasp while minimizing the possible negative impact of adding inferior grasps to the database.

For detecting similar grasps, we use the similarity score 

\begin{equation}
    \psi = \sqrt{||\mathbf{t_{\Delta}}||^{2} + \left( \lambda \theta\right)^2},
    \label{eq:similarity}
\end{equation}

where $\mathbf{t_{\Delta}}$ denotes the translational and $\theta$ the geodesic distance between two grasp poses.
We heuristically choose a scaling factor $\lambda$ of $0.01~\mathrm{m/rad}$ and a similarity threshold of $\psi = 4~\mathrm{mm}$ as a balance between avoiding repeated failures on functionally identical grasps and excluding viable neighboring candidates.
This choice translates to a tolerance of $4~\mathrm{mm}$ in translation or $23^\circ$ in rotation, respectively.

\begin{figure}[t!]
    \centering
    \includegraphics[width=\columnwidth]{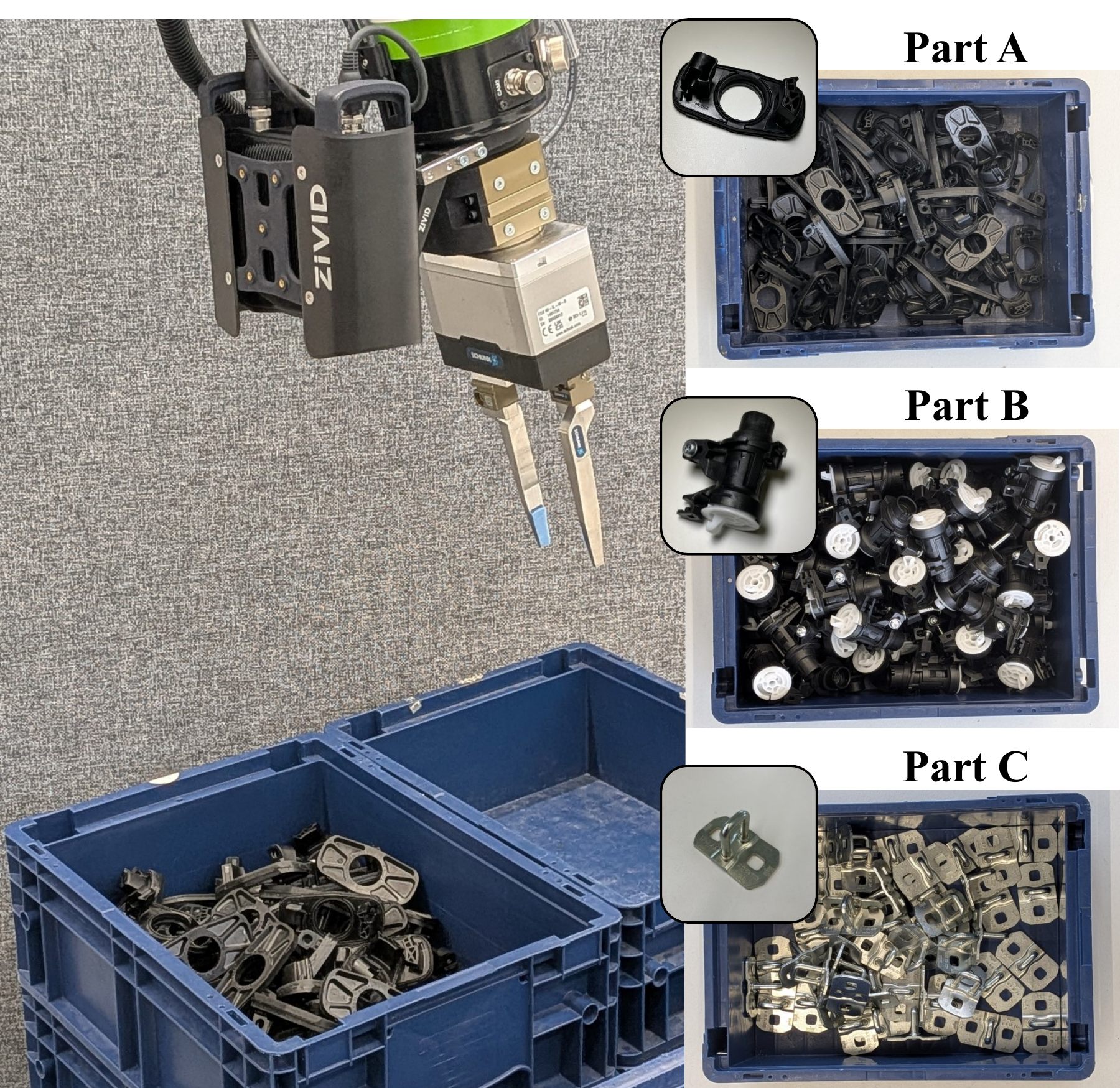}
    \caption{Bin-picking setup with an RGB-D camera and a parallel-jaw gripper with soft gripper finger pads. The right side shows the three automotive parts A, B, and C with 50 parts per bin.}
    \label{fig:setup_and_parts}
\end{figure}

The desired long-term behavior is that the system learns as many grasps as necessary to empty a bin in all situations, without generating an excessive number of additional grasps, which would negatively affect the duration of the collision check.
Over time, the generation of new grasp candidates on Level II should occur less frequently after successful grasps are discovered and added to the database.
Grasps with the highest chance of success are prioritized as the confidence interval for each grasp is narrowed down.

\section{EXPERIMENTS}
\cref{fig:setup_and_parts} shows our bin-picking setup with a Fanuc CRX-25iA cobot, equipped with a Zivid 2+ LR110 camera, providing high-quality RGB-D data.
Fingers with soft material pads are mounted on a Schunk EGK-40 parallel gripper so that a flexible form fit is created for diverse geometries.
Experiments are carried out for three automotive parts, A, B, and C, varying in appearance, size, and geometry.

We first establish the model-based and model-free baseline performance.
For both, we pick from three bins per part, starting with 50 parts per load carrier (VDA-R-KLT 4315).

For the model-based baseline, Level I is executed until a perception error causes a deadlock situation, which would require manual intervention.
If no collision-free grasp is found, we allow the system two retries by repeating the camera scan without rearranging the objects in the bin.

In the ``model-free'' case, we skip the execution of predefined grasps on Level I and instead execute Levels II to IV, predicting new grasps with CGN for each attempt.
We set the tool center point (TCP) of our gripper 5 mm inwards from the tip of the gripper for all three parts.
This then corresponds to the resulting grasp depth relative to the surface points on which CGN predicts grasp poses.

For our hybrid approach, we perform the same experiment using all four hierarchy levels and enabling the online self-learning of grasp points as well as the ranking mechanism using Wilson scores.
We run this test for five bins per part to investigate the learning across multiple bins.

\begin{figure}[t!]
    \centering
    \includegraphics[width=\columnwidth]{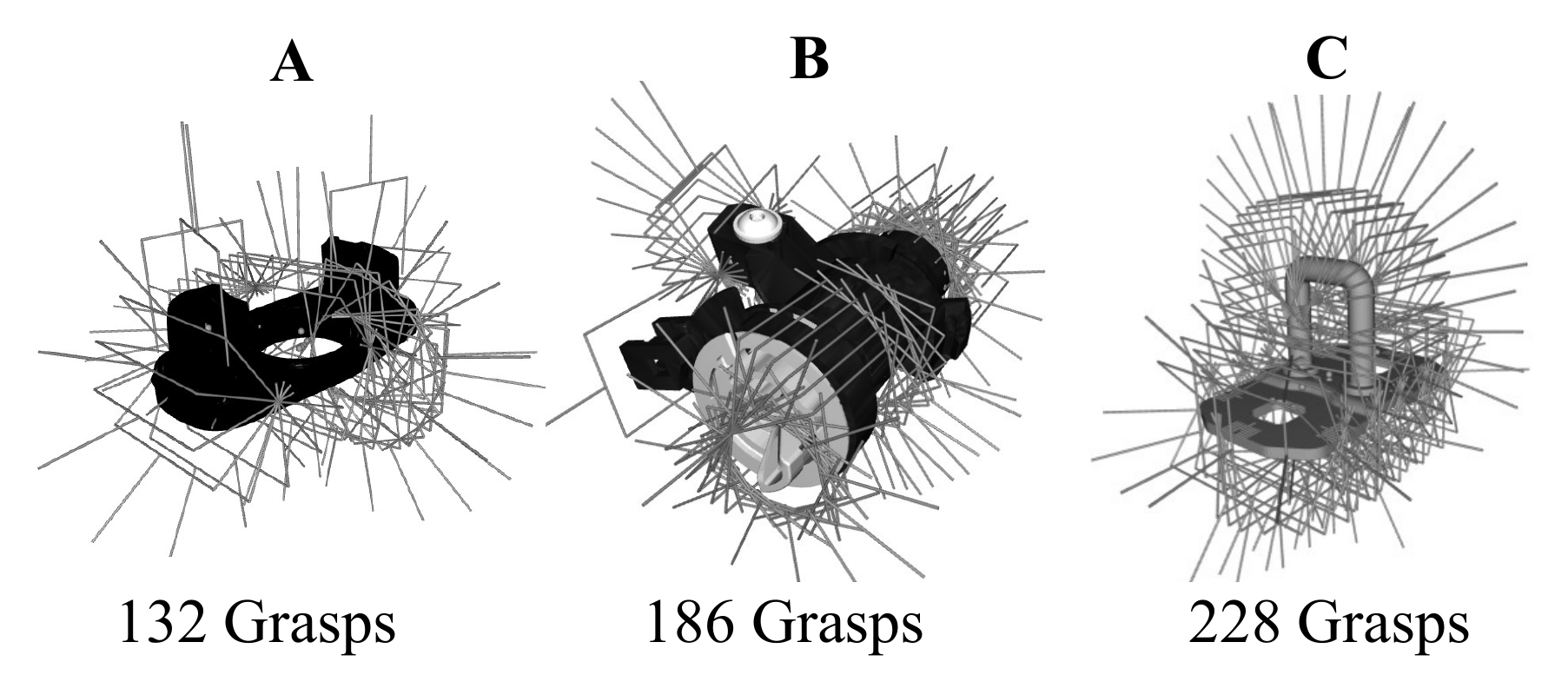}
    \caption{The grasp database of A, B, and C for manual initialization with their count, including 180° replications.}
    \label{fig:annotated_grasps}
\end{figure}

This is conducted with two initialization strategies.
A first strategy, ``Empty Init'', starts from an empty grasp database to test the effectiveness of the Wilson score ranking.
In a second experiment, ``Manual Init'', we manually initialize the grasp database to reflect a typical industrial application.
For this, an experienced user annotates grasps via a GUI, taking between 30 and 45 minutes per part. 
Again, we add a replication for each predicted grasp with a 180° flip around its z-axis to account for the symmetric gripper jaws.
The annotated grasps for the three parts and their total count, including 180° replications, are visualized in \cref{fig:annotated_grasps}.

As our performance metrics across all experiments, we calculate the bin clearance rate and the grasp success rate.
The bin clearance rate is the percentage of successfully picked parts $k$ out of the initial number of parts per bin $m$ as defined in \cref{eq:bin_clearance} \cite{Liu2024EfficientED}.

\begin{equation}
    \text{Bin Clearance Rate} = \frac{k}{m} \cdot 100
    \label{eq:bin_clearance}
\end{equation}

The grasp success rate is calculated as shown in \cref{eq:grasp_success} as the percentage of successful picks $k$ out of the total number of grasp attempts $n$ \cite{Liu2024EfficientED}.

\begin{equation}
    \text{Grasp Success Rate} = \hat{p} \cdot 100 = \frac{k}{n} \cdot 100
    \label{eq:grasp_success}
\end{equation}

\begin{table}[b!]
    \centering
    \caption{Average Bin Clearance Rates of Baselines}
    \label{tab:bin_clearance}
    \begin{tabular}{lcc}
        \toprule
        Part & Model-based ($k/m$) & Model-free ($k/m$) \\
        \midrule
        A & 33.3\% (50/150)  & 100.0\% (150/150) \\
        B & 34.7\% (52/150)  & 100.0\% (150/150) \\
        C & 84.7\% (127/150) & 100.0\% (150/150) \\
        \midrule
        All & 50.9\% (229/450) & 100.0\% (450/450) \\
        \bottomrule
    \end{tabular}
\end{table}

The outcome of each grasp attempt is detected automatically by reading the gripper width after force closure.%
A limitation of our system is that a pick of multiple entangled parts is not automatically detected yet.
This occurred in rare cases for part A and was manually corrected by placing entangled parts back in the bin.

\section{RESULTS}

The achieved bin clearance rates are shown in \cref{tab:bin_clearance}. 
For the model-based baseline, the average bin clearance across all three parts is 50.9\%.
The model-free baseline on the other hand is capable of reliably emptying all bins achieving 100\% bin clearance, as CGN predicts 2,048 grasp candidates (plus their 180° flip) for each unique situation. 

\begin{figure}[t!]
    \centering
    \includegraphics[width=\columnwidth]{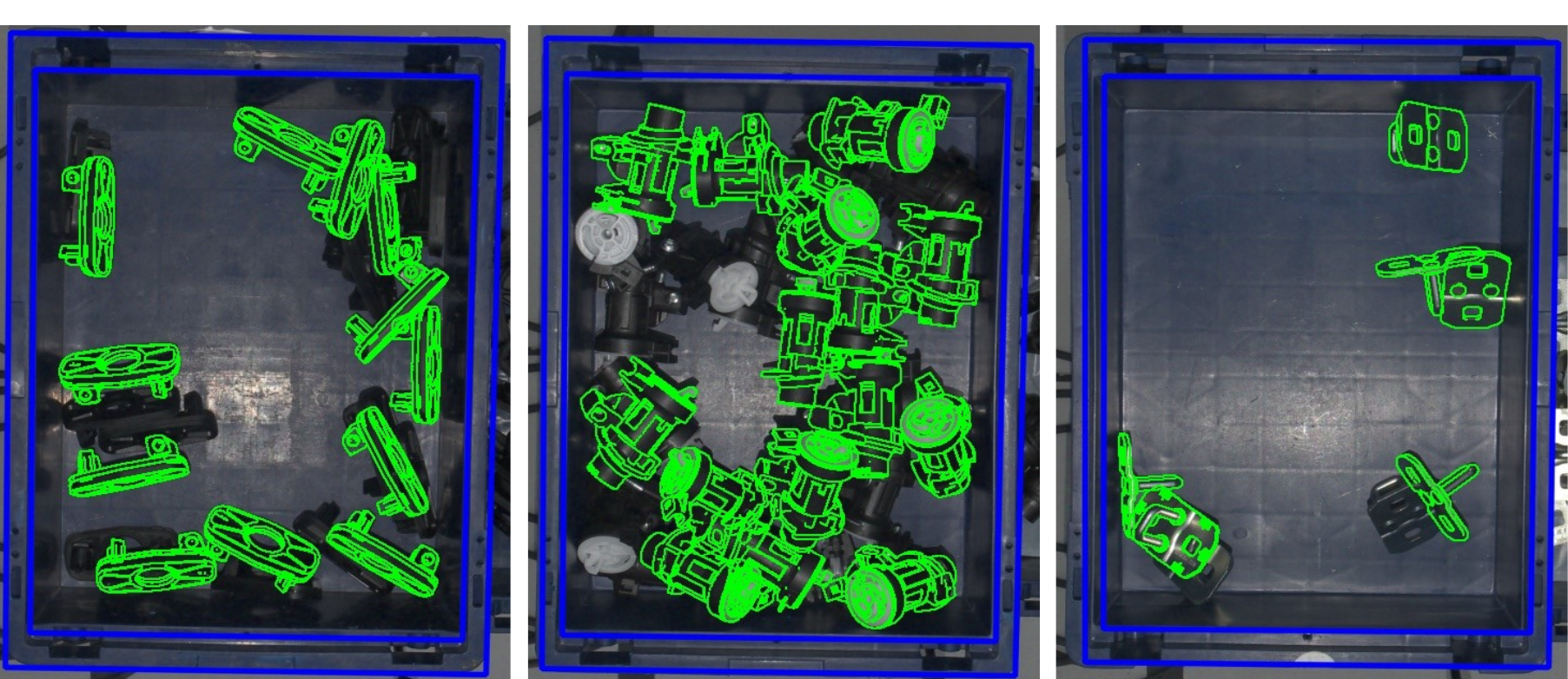}
    \caption{Bin state after the longest successful model-based run of A, B and C. Deadlocks occur as none of the predefined grasps are collision-free.}
    \label{fig:stop_conditions}
\end{figure}

The deadlock after the longest successful model-based run for each of the three parts is illustrated in \cref{fig:stop_conditions}.
It becomes evident that the main bottleneck of our system is a lack of collision-free, predefined grasp points, rather than object detection or pose estimation failures.
The results for parts A and C highlight that this issue occurs particularly for objects close to the side walls of the bin.
Part B exhibits the most complex geometry, so that deadlocks also occur when parts are still remaining in the center of the bin.

\cref{tab:grasp_success} shows the average grasp success rate of the two baselines.
As clearly shown by their discrepancy in grasp success for all three parts, the reliable bin clearance of the model-free system comes at the cost of a drastically reduced success rate.
The average across all parts is only 45.6\%, whereas the model-based baseline achieves a near perfect score of 99.1\%.
We observe that CGN predicts many unstable grasps, particularly for part B, as the necessary grasp depth of the TCP relative to the contact point varies significantly for different high curvature areas but is not predicted by the model architecture.

\begin{table}[b!]
    \centering
    \caption{Average Grasp Success Rates of Baselines}
    \label{tab:grasp_success}
    \begin{tabular}{lcc}
        \toprule
        Part & Model-based ($k/n$) & Model-free ($k/n$)\\
        \midrule
        A & 96.1\% (50/52)   & 71.4\% (150/210) \\
        B & 100.0\% (52/52)  & 29.8\% (150/503) \\
        C & 100.0\% (127/127) & 55.0\% (150/273) \\
        \midrule
        All & 99.1\% (229/231) & 45.6\% (450/986) \\
        \bottomrule
    \end{tabular}
\end{table}

\begin{figure}[t!]
    \centering
    \includegraphics[width=\columnwidth]{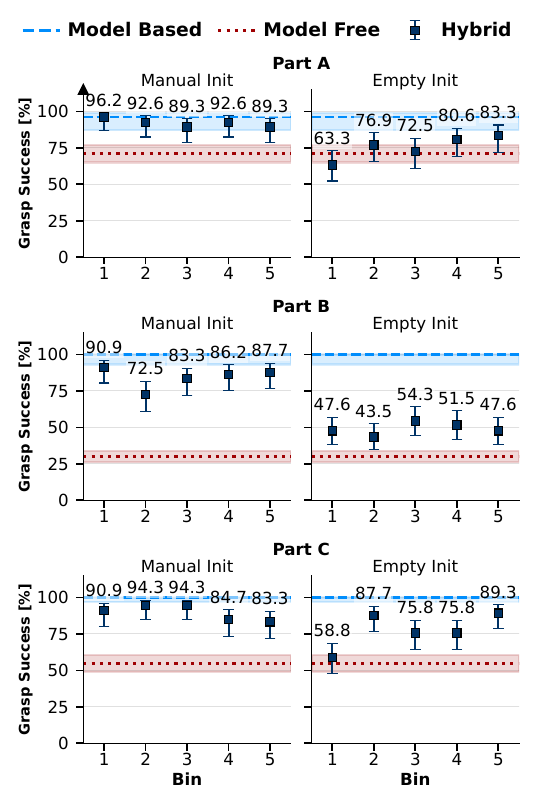}
    \caption{The grasp success rate for A, B, and C over five bins with manual and empty initialization. Each entry shows the 95\% confidence interval calculated using \cref{eq:wilson}.}
    \label{fig:plot_learning}
\end{figure}

Crucially, our hierarchical hybrid approach maintains the 100\% bin clearance rate of the model-free baseline for all 30 bins across the experiments.
Throughout the 15 bins with manual initialization, deadlocks were overcome successfully 18 times through grasp point discovery on Level II, once through model-free grasping on Level III, and three times by the ``unknown'' grasping on Level IV.
In some cases, even unsuccessful grasps resolved deadlock situations by repositioning parts into more favorable poses.

These findings indicate that object detection and 6D pose estimation are already comparatively robust for the analyzed objects.
Consequently, the primary bottleneck to achieve continuous industrial operation lies in discovering high-quality, collision-free grasps.

The grasp success rates of the hybrid approach are visualized in \cref{fig:plot_learning} for the two initialization strategies over five bins each, along with their 95\% Wilson confidence interval.
As the plot illustrates, the grasp success rate after manual initialization remains consistently high, achieving an average of 91.9\% for A, 83.6\% for B, and 89.3\% for C.
It must be noted that the model-based baseline stops without executing grasps in the most challenging configurations, especially towards the end of a bin, contributing to the observed discrepancy in grasp success.

To estimate the excess failures caused by exploring learned grasps when predefined ones are still available on Level I ("pre-deadlock"), we calculate a hypothetical GSR penalty

\begin{equation}
    \Delta \text{GSR} = \frac{e_{\text{learned}} - (n_{\text{learned}} \cdot e_{\text{manual}} / n_{\text{manual}})}{n_{\text{total}}}
    \label{eq:GSR_penalty}
\end{equation}

where $e_{manual}$ and $e_{learned}$ represent the errors for predefined and discovered grasps, respectively.
The resulting $\Delta \text{GSR}$ is 2.6\% (A), 1.2\% (B), and 0.0\% (C). 
This slight initial drop-off is expected to diminish over time as the ranking algorithm further prioritizes the most robust grasps.

With respect to cycle time, Levels II–IV introduce a 1.01\,s overhead for grasp generation and 1.36\,s for collision checking the 4,096 predicted grasps (with 180° replications).
Given the minimal collision checking cost of approximately 0.3\,ms per grasp, extending the databases (by 22, 10, and 12 grasps for A, B, and C throughout the experiments) has a negligible impact on the cycle time of Level I.

For empty initialization, our results show a significant improvement over the model-free baseline for all three parts.
However, the final performance varies significantly between the different objects.
For part B, we observe the same part-specific challenges for CGN as with the model-free baseline.

\begin{figure}[t!]
    \centering
    \includegraphics[width=\columnwidth]{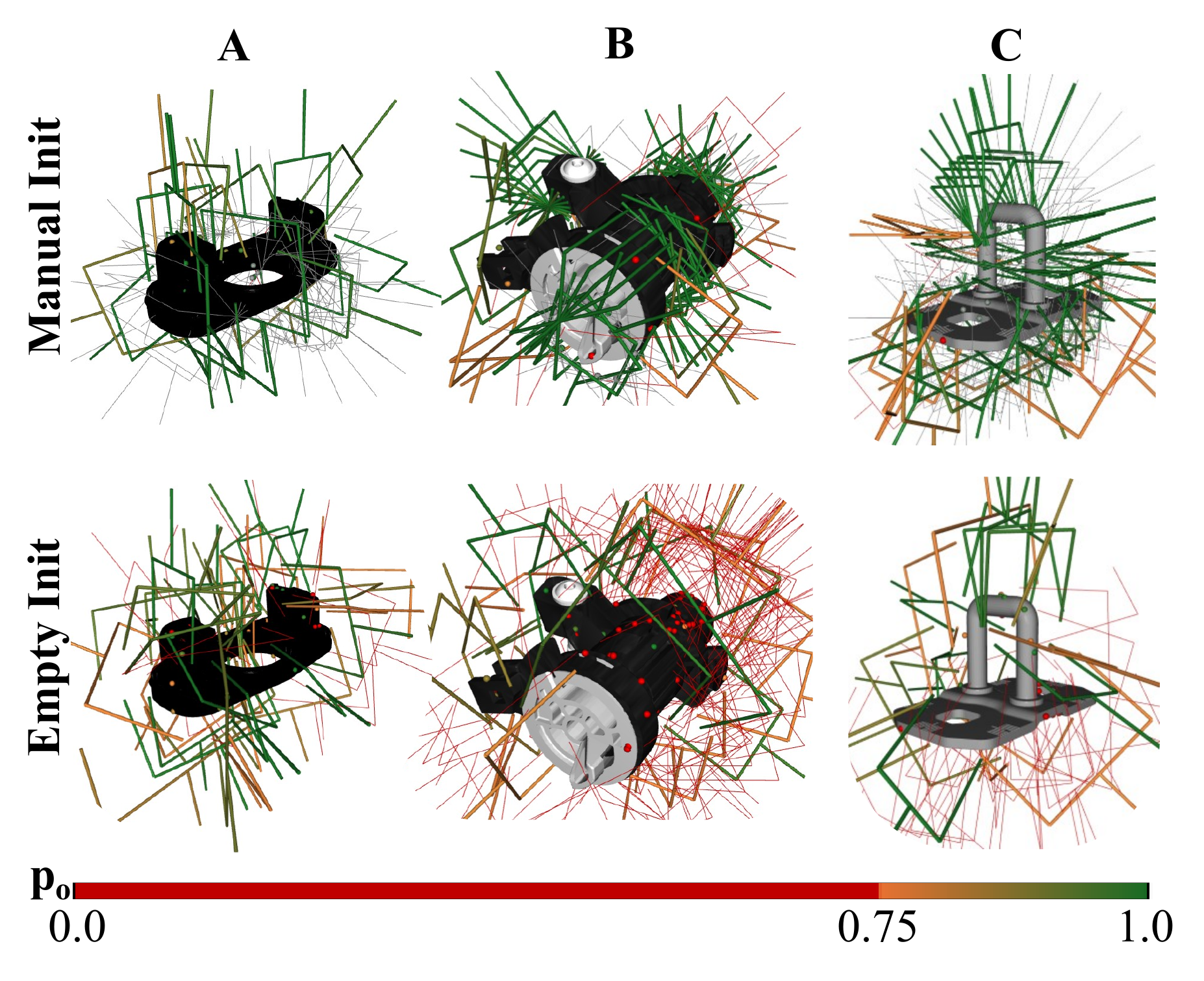}
    \caption{The database of grasp points for A, B, and C after the fifth bin with manual and empty initialization along with the respective $p_o$ values.}
    \label{fig:database_after_learning}
\end{figure}

The grasp databases of parts A, B, and C after the fifth bin are visualized for both initialization strategies in \cref{fig:database_after_learning}.
Predefined grasps which were not executed are shown in gray.
Manually predefined grasps were executed consistently with high success.
The model-free exploration added some additional---sometimes non-intuitive---grasps, including less robust grasps which were situationally necessary to resolve edge cases.
For empty initialization, the persistent issues for the grasp depth of part B become apparent.
For parts A and C, the plot showcases the system's ability to successfully discover and prioritize reliable grasp poses.

\section{CONCLUSION}
The results validate our initial assumption that model-free grasp generation can meaningfully complement the strengths and weaknesses of model-based bin-picking.
Our hybrid approach allows the system to autonomously resolve deadlocks, achieving 100\% bin clearance without human intervention or manual fine-tuning.
At the same time, it achieves high grasp success and repeatability, especially after manual initialization, due to the model-based backbone and its 6D pose estimation.

The approach is independent of the concrete models forming the model-based backbone on Level I.
With that, it can enhance existing bin-picking systems and contribute to significantly improved autonomy and economic value.
We see the largest potential in HMLV manufacturing, where reduced manual commissioning effort for new parts is crucial.

For future work, we see strong potential in few- or zero-shot bin-picking.
Our hierarchical pipeline could act as a robust safety net to handle a potential increase in perception errors compared to state-of-the-art supervised models.
To create a fully zero-shot, plug-and-produce system, CAD-based initialization of the grasp database could be employed, following the work of Kleeberger et al.~\cite{kleeberger2021automatic}, instead of the manual initialization used for maximum performance in our current implementation.

Furthermore, large potential lies in optimizing the model-free grasp generation.
Fine-tuning networks like CGN for cluttered industrial parts, or employing an alternative model architecture which predicts point-specific grasp depth, could enhance the grasp success rates on Levels II-IV and, with that, the overall system performance.
Alongside this, extended experiments and ablations are necessary to analyze the long-term system behavior and to determine the optimal choice of parameters for the learning mechanism.

\bibliographystyle{IEEEtran}

\bibliography{IROS2026_references}

\begin{thebibliography}{10}
\providecommand{\url}[1]{#1}
\csname url@rmstyle\endcsname
\providecommand{\newblock}{\relax}
\providecommand{\bibinfo}[2]{#2}
\providecommand\BIBentrySTDinterwordspacing{\spaceskip=0pt\relax}
\providecommand\BIBentryALTinterwordstretchfactor{4}
\providecommand\BIBentryALTinterwordspacing{\spaceskip=\fontdimen2\font plus
\BIBentryALTinterwordstretchfactor\fontdimen3\font minus
  \fontdimen4\font\relax}
\providecommand\BIBforeignlanguage[2]{{%
\expandafter\ifx\csname l@#1\endcsname\relax
\typeout{** WARNING: IEEEtran.bst: No hyphenation pattern has been}%
\typeout{** loaded for the language `#1'. Using the pattern for}%
\typeout{** the default language instead.}%
\else
\language=\csname l@#1\endcsname
\fi
#2}}

\bibitem{huang2025xyz}
J.~Huang \emph{et~al.}, ``{XYZ-IBD}: A high-precision bin-picking dataset for
  object {6D} pose estimation capturing real-world industrial complexity,''
  \emph{arXiv preprint arXiv:2506.00599}, 2025.

\bibitem{Liu2024EfficientED}
Y.~Liu \emph{et~al.}, ``Efficient end-to-end detection of {6-DoF} grasps for
  robotic bin picking,'' in \emph{2024 IEEE International Conference on
  Robotics and Automation (ICRA)}.\hskip 1em plus 0.5em minus 0.4em\relax IEEE,
  2024, pp. 5427--5433.

\bibitem{spenrath2022heuristisches}
F.~Spenrath, ``{Heuristisches Suchverfahren f{\"u}r die effiziente Planung zum
  Greifen ungeordnet gelagerter Werkst{\"u}cke mit Industrierobotern},'' Ph.D.
  dissertation, Fraunhofer-Institut f{\"u}r Produktionstechnik und
  Automatisierung IPA, Stuttgart, 2022.

\bibitem{sarna2023impact}
M.~Sarna, A.~L{\"u}der, A.~Tegtmeier, J.~Weist, and S.~Espericueta, ``Impact of
  modified bin picking container,'' in \emph{2023 3rd International Conference
  on Electrical, Computer, Communications and Mechatronics Engineering
  (ICECCME)}.\hskip 1em plus 0.5em minus 0.4em\relax IEEE, 2023, pp. 1--7.

\bibitem{Boschetti2023}
G.~Boschetti, T.~Sinico, and A.~Trevisani, ``Improving robotic bin-picking
  performances through human--robot collaboration,'' \emph{Applied Sciences},
  vol.~13, no.~9, p. 5429, 2023.

\bibitem{Fujita2019}
M.~Fujita \emph{et~al.}, ``Bin-picking robot using a multi-gripper switching
  strategy based on object sparseness,'' in \emph{2019 IEEE 15th International
  Conference on Automation Science and Engineering (CASE)}.\hskip 1em plus
  0.5em minus 0.4em\relax IEEE, 2019, pp. 1540--1547.

\bibitem{pantano2024simplifying}
M.~Pantano \emph{et~al.}, ``Simplifying robot grasping in manufacturing with a
  teaching approach based on a novel user grasp metric,'' \emph{Procedia
  Computer Science}, vol. 232, pp. 1961--1971, 2024.

\bibitem{kleeberger2021automatic}
K.~Kleeberger, F.~Roth, R.~Bormann, and M.~F. Huber, ``Automatic grasp pose
  generation for parallel jaw grippers,'' in \emph{International Conference on
  Intelligent Autonomous Systems}.\hskip 1em plus 0.5em minus 0.4em\relax
  Springer, 2021, pp. 594--607.

\bibitem{fu2024low}
X.~Fu, L.~Miao, Y.~Ohnishi, Y.~Hasegawa, and M.~Suwa, ``A low-cost, high-speed,
  and robust bin picking system for factory automation enabled by a non-stop,
  multi-view, and active vision scheme,'' in \emph{2024 IEEE/RSJ International
  Conference on Intelligent Robots and Systems (IROS)}.\hskip 1em plus 0.5em
  minus 0.4em\relax IEEE, 2024, pp. 11\,566--11\,573.

\bibitem{sundermeyer2021contact}
M.~Sundermeyer, A.~Mousavian, R.~Triebel, and D.~Fox, ``{Contact-GraspNet}:
  Efficient {6-DoF} grasp generation in cluttered scenes,'' in \emph{2021 IEEE
  international conference on robotics and automation (ICRA)}.\hskip 1em plus
  0.5em minus 0.4em\relax IEEE, 2021, pp. 13\,438--13\,444.

\bibitem{back2025graspclutter6d}
S.~Back \emph{et~al.}, ``{GraspClutter6D}: A large-scale real-world dataset for
  robust perception and grasping in cluttered scenes,'' \emph{IEEE Robotics and
  Automation Letters}, 2025.

\bibitem{khalid2021automatic}
M.~U. Khalid \emph{et~al.}, ``Automatic grasp generation for vacuum grippers
  for random bin picking,'' in \emph{Advances in Automotive Production
  Technology--Theory and Application: Stuttgart Conference on Automotive
  Production (SCAP2020)}.\hskip 1em plus 0.5em minus 0.4em\relax Springer,
  2021, pp. 247--255.

\bibitem{cordeiro2025review}
A.~Cordeiro, L.~F. Rocha, J.~Boaventura-Cunha, D.~Figueiredo, and J.~P. Souza,
  ``A review of visual perception for robotic bin-picking,'' \emph{Robotics and
  Autonomous Systems}, p. 105236, 2025.

\bibitem{kleeberger2020transferring}
K.~Kleeberger \emph{et~al.}, ``Transferring experience from simulation to the
  real world for precise pick-and-place tasks in highly cluttered scenes,'' in
  \emph{2020 IEEE/RSJ International Conference on Intelligent Robots and
  Systems (IROS)}.\hskip 1em plus 0.5em minus 0.4em\relax IEEE, 2020, pp.
  9681--9688.

\bibitem{kleeberger2020survey}
K.~Kleeberger, R.~Bormann, W.~Kraus, and M.~F. Huber, ``A survey on
  learning-based robotic grasping,'' \emph{Current Robotics Reports}, vol.~1,
  no.~4, pp. 239--249, 2020.

\bibitem{moosmann2020increasing}
M.~Moosmann \emph{et~al.}, ``Increasing the robustness of random bin picking by
  avoiding grasps of entangled workpieces,'' \emph{Procedia CIRP}, vol.~93, pp.
  1212--1217, 2020.

\bibitem{huang2025lessons}
Z.~Huang \emph{et~al.}, ``Lessons and winning solutions in industrial object
  detection and pose estimation from the 2025 bin-picking perception
  challenge,'' in \emph{Proceedings of the IEEE/CVF International Conference on
  Computer Vision}, 2025, pp. 2408--2414.

\bibitem{bop_challenges_website}
M.~Sundermeyer \emph{et~al.}, ``{BOP} challenge: Benchmark for {6D} object pose
  estimation,'' \url{https://bop.felk.cvut.cz/challenges/}, 2025, accessed:
  2026-02-06.

\bibitem{cao20236impose}
H.~Cao, L.~Dirnberger, D.~Bernardini, C.~Piazza, and M.~Caccamo, ``{6IMPOSE}:
  Bridging the reality gap in {6D} pose estimation for robotic grasping,''
  \emph{Frontiers in Robotics and AI}, vol.~10, p. 1176492, 2023.

\bibitem{chen2025zerobp}
J.~Chen, Z.~Zhou, X.~Li, Y.~Zheng, T.~Bao, and Z.~He, ``{ZeroBP}: Learning
  position-aware correspondence for zero-shot {6D} pose estimation in
  bin-picking,'' in \emph{2025 IEEE International Conference on Robotics and
  Automation (ICRA)}.\hskip 1em plus 0.5em minus 0.4em\relax IEEE, 2025, pp.
  12\,266--12\,272.

\bibitem{hagelskjaer2025good}
F.~Hagelskj{\ae}r, ``Good grasps only: A data engine for self-supervised
  fine-tuning of pose estimation using grasp poses for verification,'' in
  \emph{2025 IEEE/SICE International Symposium on System Integration
  (SII)}.\hskip 1em plus 0.5em minus 0.4em\relax IEEE, 2025, pp. 957--964.

\bibitem{murali2025graspgen}
A.~Murali \emph{et~al.}, ``{GraspGen}: A diffusion-based framework for {6-DoF}
  grasping with on-generator training,'' \emph{arXiv preprint
  arXiv:2507.13097}, 2025.

\bibitem{mahler2016dex}
J.~Mahler \emph{et~al.}, ``{Dex-Net 1.0}: A cloud-based network of {3D} objects
  for robust grasp planning using a multi-armed bandit model with correlated
  rewards,'' in \emph{2016 IEEE international conference on robotics and
  automation (ICRA)}.\hskip 1em plus 0.5em minus 0.4em\relax IEEE, 2016, pp.
  1957--1964.

\bibitem{Sanders2020}
K.~Sanders, M.~Danielczuk, J.~Mahler, A.~Tanwani, and K.~Goldberg, ``Non-markov
  policies to reduce sequential failures in robot bin picking,'' in \emph{2020
  IEEE 16th International Conference on Automation Science and Engineering
  (CASE)}.\hskip 1em plus 0.5em minus 0.4em\relax IEEE, 2020, pp. 1141--1148.

\bibitem{pinto2016supersizing}
L.~Pinto and A.~Gupta, ``Supersizing self-supervision: Learning to grasp from
  50k tries and 700 robot hours,'' in \emph{2016 IEEE international conference
  on robotics and automation (ICRA)}.\hskip 1em plus 0.5em minus 0.4em\relax
  IEEE, 2016, pp. 3406--3413.

\bibitem{levine2018learning}
S.~Levine, P.~Pastor, A.~Krizhevsky, J.~Ibarz, and D.~Quillen, ``Learning
  hand-eye coordination for robotic grasping with deep learning and large-scale
  data collection,'' \emph{The International journal of robotics research},
  vol.~37, no. 4-5, pp. 421--436, 2018.

\bibitem{monorchio2018learning}
L.~Monorchio, D.~Evangelista, M.~Imperoli, and A.~Pretto, ``Learning from
  successes and failures to grasp objects with a vacuum gripper,'' in
  \emph{IEEE/RSJ IROS Workshop on Task-Informed Grasping for Rigid and
  Deformable Object Manipulation}, 2018.

\bibitem{patten2020dgcm}
T.~Patten, K.~Park, and M.~Vincze, ``{DGCM-Net}: Dense geometrical
  correspondence matching network for incremental experience-based robotic
  grasping,'' \emph{Frontiers in Robotics and AI}, vol.~7, p. 120, 2020.

\bibitem{komoda2024hybrid}
K.~Komoda \emph{et~al.}, ``{Hybrid-AI} grasp planning system that integrates
  rule-based and {DNN-based} methods for throughput improvement of picking
  robots,'' in \emph{2024 IEEE International Conference on Advanced Intelligent
  Mechatronics (AIM)}.\hskip 1em plus 0.5em minus 0.4em\relax IEEE, 2024, pp.
  691--696.

\bibitem{roboception_cadmatch}
{Roboception GmbH}, ``{CADMatch Machine Tending Software},''
  \url{https://roboception.com/cadmatch-machine-tending-software/}, 2026,
  accessed: 2026-02-15.

\bibitem{murray2017mathematical}
R.~M. Murray, Z.~Li, and S.~S. Sastry, \emph{A mathematical introduction to
  robotic manipulation}.\hskip 1em plus 0.5em minus 0.4em\relax CRC press,
  2017.

\bibitem{Wilson01061927}
E.~B. Wilson, ``Probable inference, the law of succession, and statistical
  inference,'' \emph{Journal of the American Statistical Association}, vol.~22,
  no. 158, pp. 209--212, 1927.

\bibitem{auer2002finite}
P.~Auer, N.~Cesa-Bianchi, and P.~Fischer, ``Finite-time analysis of the
  multiarmed bandit problem,'' \emph{Machine learning}, vol.~47, no.~2, pp.
  235--256, 2002.

\bibitem{daniel2018tutorial}
D.~J. Russo, B.~Van~Roy, A.~Kazerouni, I.~Osband, and Z.~Wen, ``A tutorial on
  thompson sampling,'' \emph{Foundations and Trends{\textregistered} in Machine
  Learning}, vol.~11, no.~1, pp. 1--99, 2018.

\end{thebibliography}

\end{document}